\pdfoutput=1
\documentclass{article} 

\usepackage{microtype}
\usepackage{graphicx}
\usepackage{subcaption}
\usepackage{booktabs} 
\usepackage{threeparttable}

\usepackage{algorithmicx}
\usepackage{algorithm}
\usepackage[noend]{algpseudocode}
\usepackage[utf8]{inputenc} 

\usepackage{times}

\usepackage[accepted]{icml2019}

\icmltitlerunning{Uncertainty-sensitive Learning and Planning with Ensembles}

\usepackage{amsmath,amsfonts,bm}

\def\eqref#1{equation~\ref{#1}}

\def\1{\bm{1}}

\DeclareMathAlphabet{\mathsfit}{\encodingdefault}{\sfdefault}{m}{sl}
\SetMathAlphabet{\mathsfit}{bold}{\encodingdefault}{\sfdefault}{bx}{n}

\DeclareMathOperator*{\argmax}{arg\,max}
\DeclareMathOperator*{\argmin}{arg\,min}

\usepackage{url}

\usepackage[utf8]{inputenc} 
\usepackage[T1]{fontenc}    
\usepackage{hyperref}
\usepackage{url}            
\usepackage{booktabs}       
\usepackage{amsfonts}       
\usepackage{amsmath}
\usepackage{nicefrac}       
\usepackage{microtype}      

\usepackage{todonotes}

\usepackage[font={footnotesize,it}]{caption}

\usetikzlibrary{shapes.geometric,backgrounds,
  positioning-plus,node-families,calc}
\tikzset{
  basic box/.style = {
    shape = rectangle,
    align = center,
    draw  = #1,
    fill  = #1!25,
    rounded corners},
  header node/.style = {
    Minimum Width = header nodes,
    font          = \strut\Large\ttfamily,
    text depth    = +0pt,
    fill          = white,
    draw},
  header/.style = {%
    inner ysep = +1.5em,
    append after command = {
      \pgfextra{\let\TikZlastnode\tikzlastnode}
      node [header node] (header-\TikZlastnode) at (\TikZlastnode.north) {#1}
      node [span = (\TikZlastnode)(header-\TikZlastnode)]
        at (fit bounding box) (h-\TikZlastnode) {}
    }
  },
  hv/.style = {to path = {-|(\tikztotarget)\tikztonodes}},
  vh/.style = {to path = {|-(\tikztotarget)\tikztonodes}},
  fat blue line/.style = {ultra thick, blue}
}

\usetikzlibrary{arrows,decorations.pathmorphing,backgrounds,positioning}

\definecolor{echoreg}{HTML}{2cb1e1}
\definecolor{olivegreen}{rgb}{0,0.6,0}
\definecolor{mymauve}{rgb}{0.58,0,0.82}

\usepackage{etoolbox}

\newtoggle{redraw}
\newtoggle{redraw2}

\usepackage{wrapfig}

\tikzset{%
pics/cube/.style args={#1/#2/#3/#4}{code={%
	\begin{scope}[line width=#4mm]
	\begin{scope}
	\clip (-#1,-#2,0) -- (#1,-#2,0) -- (#1,#2,0) -- (-#1,#2,0) -- cycle;
	\filldraw (-#1,-#2,0) -- (#1,-#2,0) -- (#1,#2,0) -- (-#1,#2,0) -- cycle;
	\end{scope}
\iftoggle{redraw}{%
}{%
	\begin{scope}
	\clip (-#1,-#2,0) -- (-#1-#3,-#2,-#3) -- (-#1-#3,#2,-#3) -- (-#1,#2,0) -- cycle;
	\filldraw (-#1,-#2,0) -- (-#1-#3,-#2,-#3) -- (-#1-#3,#2,-#3) -- (-#1,#2,0) -- cycle;
	\end{scope}
}
\iftoggle{redraw2}{%
}{
	\begin{scope}
	\clip (-#1,#2,0) -- (-#1-#3,#2,-#3) -- (#1-#3,#2,-#3) -- (#1,#2,0) -- cycle;
	\filldraw (-#1,#2,0) -- (-#1-#3,#2,-#3) -- (#1-#3,#2,-#3) -- (#1,#2,0) -- cycle;
	\end{scope}
}
	\node[inner sep=0] (-A) at (-#1-#3*0.5, 0, -#3*0.5) {};
	\node[inner sep=0] (-B) at (#1-#3*0.5, 0, -#3*0.5) {};
	
	\coordinate (-V) at (#1, #2);
	\coordinate (-W) at (#1, -#2);
	\end{scope}
}}}

\usepackage{amsmath}

\definecolor{myred}{cmyk}{0,0.9,0.9,0.1}
\definecolor{myblue}{cmyk}{0,0.3,0.9,0.1}

\makeatletter
\newcommand{\removelatexerror}{\let\@latex@error\@gobble}
\makeatother
\usepackage{comment}

\usepackage{pdflscape}

\begin{document}

\twocolumn[
\icmltitle{Uncertainty-sensitive Learning and Planning with Ensembles}

\icmlsetsymbol{equal}{*}

\begin{icmlauthorlist}
\icmlauthor{Piotr Miłoś}{equal,impan,ds}
\icmlauthor{Łukasz Kuciński}{equal,impan}
\icmlauthor{Konrad Czechowski}{equal,uw}
\icmlauthor{Piotr Kozakowski}{uw}
\icmlauthor{Maciej Klimek}{uw}

\end{icmlauthorlist}

\icmlaffiliation{uw}{Faculty of Mathematics, Informatics and Mechanics, University of Warsaw, Warsaw, Poland}
\icmlaffiliation{impan}{Institute of Mathematics Polish Academy of Sciences, Warsaw, Poland}
\icmlaffiliation{ds}{deepsense.ai, Warsaw, Poland}

\icmlcorrespondingauthor{Łukasz Kuciński}{lukasz.kucinski@impan.pl}

\icmlcorrespondingauthor{Konrad Czechowski}{k.czechowski@mimuw.edu.pl}

\icmlkeywords{reinforcement learning, mcts, uncertainty}

\vskip 0.3in
]

\printAffiliationsAndNotice{\icmlEqualContribution}

\begin{abstract}
We propose a reinforcement learning framework for discrete environments in which an agent makes both strategic and tactical decisions. The former manifests itself through the use of value
function, while the latter is powered by a tree search planner. These tools complement each other. 
The planning module performs a local \mbox{what-if} analysis, which allows avoiding tactical pitfalls and boosts backups of the value function.
The value function, being global in nature, compensates for the inherent locality of the planner. In order to further solidify this synergy, we introduce an exploration mechanism with two  components: uncertainty modelling and risk measurement. To model the uncertainty, we use value function ensembles, and to reflect risk, we use several functionals that summarize the uncertainty implied by the ensemble. We show that our method performs well on hard exploration environments: Deep-sea, toy Montezuma’s Revenge, and Sokoban. In all the cases, we obtain speed-up in learning and boost in performance.
\end{abstract}
\section{Introduction}\label{sec:introduction}

The model-free and model-based approaches to reinforcement learning (RL) have complementary sets of strengths and weaknesses. While the former offers good asymptotic performance, it suffers from inferior sample complexity. 
In contrast, the latter usually needs significantly less training samples, but often fails to achieve state-of-the-art results on complex tasks (which is primarily attributed to the model's imperfections). The interplay of model-based and model-free approaches in RL has received a lot of research attention. This led, for example, to strong AI systems like \citet{Silver2017, Silver2018} or more recently to \citet{Lowrey2018}, which is closely related to our work.

When dealing with challenging RL domains, it is helpful to address strategic and tactical decision-making. These two perspectives complement each other: the strategic perspective is global, static, and often approximate, while the tactical perspective is local, dynamic, and exact. A neural network value function can be considered as an implementation of the former, while a planner as an example of the latter. Indeed, neural network value functions provide noisy estimates (approximate) of values (static) to every state (global). Conversely, planning provides high-quality control which, starting from a given state (local), generates actions (dynamic) that are temporally coherent and result in better executed trajectories (exact).

In this paper we propose a framework combining the aforementioned components into a single system.
We test our approach on the sparse reward variants of the following environments: Sokoban, a classic logical puzzle known for its combinatorial complexity (in fact, answering the question of whether a Sokoban level is solvable is NP-hard, see e.g. \citet{dor1999sokoban}) and a recent benchmark for RL, ChainEnv, a seemingly impossible task referred to as a 'hay in a needle-stack' problem (see \citet{Osband2018a}), and Toy Montezuma’s Revenge, environment notoriously known for its exploration difficulty (see \citet{moczulskimontezuma2019}).

Put differently, we consider a situation of an agent with limited memory and computational resources, being dropped into a complex and diverse environment.
We assume that solving for an optimal trajectory is out of reach, and a limited depth planner has to be used. 
Plugging the value function into the planner provides guided heuristics in the local search, shortens the search horizon, and thus makes the search computationally efficient. For complex problems, this setup has to be supplemented by an exploration scheme. We develop a scheme based on modelling uncertainty of the value function approximation using ensembles. The uncertainty is quantified by a risk measure, which is then utilized by a planner to drive exploration.

The main contribution of this work is showing how recent progress in AI can be brought together to improve planning, value function learning, and exploration, in a way that forms robust algorithms for solving challenging reinforcement learning environments. In particular: 
\begin{enumerate}
\item For uncertainty modeling, we assume the point of view of \citet{Osband2018a}, 
which uses ensembles to approximate posterior distribution. 

\item In the spirit of \citet{Lowrey2018}, we incorporate risk measures to guide exploration.  

\item For the planner, we base on AlphaZero Monte-Carlo Tree Search, see \citet{Silver2017}.

\item In the value function training protocol, we introduce several improvements, including a version of prioritized replay buffer and  hindsight (\citet{hind19}). We found it beneficial to calculate targets for value function learning using the planner search history.
\end{enumerate}

The rest of the paper is organized as follows. In the next subsection we provide an overview of related work. In Section \ref{sec:method} we present and discuss our method in detail.
Experiments are gathered in Section \ref{sec:experiments}. Section \ref{sec:conclusions} concludes the paper. Some details concerning the algorithm's pseudo-code, neural network architecture and training, and ablations can be found in the Appendix.  
We provide source code to our work \url{https://github.com/learningandplanningICLR/learningandplanning} and a dedicated website \url{https://sites.google.com/view/learn-and-plan-with-ensembles} with more details and movies.

\subsection{Related work}
The ideas of mixing model-based and model-free learning were perhaps first stated explicitly in \citet{Sutton90}. Many approaches followed. More recently, in the groundbreaking series of papers \citet{Silver2017, Silver2018} culminating in AlphaZero, the authors have developed an elaborate system that plans and performs model-free training to master the game of Go (and others). Similar ideas were also studied in \citet{Anthony2017}. A recent paper \citet{Schrittwieser2019} presents impressive results of joint model learning and planning in the latent space.

Perhaps, the work which is closest to ours is \citet{Lowrey2018}.
It is argued in the paper that an agent with limited computational resources in a complex environment needs both to plan and learn from the incoming stream of experience. Importantly, the value function in \citet{Lowrey2018} is modeled by an ensemble of value functions. The risk measure used to combine them is given by the 'log-sum-exp formula',  \citet[equation 6]{Lowrey2018}. The authors show experimentally that this approach leads to improvements in various continuous control tasks, including Humanoid. Similar line of research was followed in recent \citet{Lu2019}. In our work, we deal with a discrete action setting, which enforces a different  planning module (here MCTS-based). Moreover, we treat a more diverse class of risk measures (see Section \ref{sec:method}).

Constructing neural network models that would incorporate uncertainty in a principled Bayesian way has proven to be challenging and remains an open problem. A promising new results using ensembles include \citet{Osband2018a, Osband2018b, Lakshminarayanan2017}. Including uncertainty in RL dates at least to \citet{Strens00} who proposes learning a model of an MDP in a Bayesian framework.  Practical initiations have been proposed e.g. \citet{Janz2019}. In our work, we use an ensemble approach of \citet{Osband2018a, Osband2018b}, which can also be viewed in this setting; see discussion in \citet[Section 2]{Janz2019}. Another interesting idea has been presented in \citet{ODonoghue2018}, to relate uncertainty at any time-step to the expected uncertainties at subsequent time-steps. \citet{Moerland2017} proposes to disentangle epistemic and aleatoric uncertainties, using Bayesian drop-out to treat the former and Gaussian distributions for modelling the latter. Ensembles of models were successfully used to improve model-based RL training, see \citet{Kurutach2018, Chua2018}, and the references therein. We note that RL practitioners willingly use unprincipled ensemble methods. For example, in a recent competition \citet{Kidzinski2019} aimed to train an agent able to use a prosthetic leg, four top-ten solutions used some ensemble-based techniques.

Another work similar to ours is \citet{Guo2014}, in which the authors use MCTS in the role of an `expert' from which a neural policy is learnt using the DAgger algorithm, \citet{dagger}. The basic difference is that \citet{Guo2014} uses a classical MCTS without value function nor ensembles.

Many works aim to build planning and learning into neural network architectures, see e.g. \citet{Oh2017, Farquhar2017}.
\citet{Kaiser19}, a recent work on model-based Atari, has shown the possibility of sample efficient reinforcement learning with an explicit visual model. \citet{GuLSL16} uses model-based methods at the initial phase of training and model-free methods during `fine-tuning'. Furthermore, there is a body of work that attempts to learn the planning module, see \citet{Pascanu2017, RacaniereWRBGRB17, Guez19}.

Finally, our paper is related to research focusing on study of exploration. 
Fundamental results in this area concern the multi-arm bandits problem, see \citet{lattimore2018bandit} and the references therein. 
Methods developed in this area have been successfully applied in planning algorithms, see \citet{kocsis2006improved} and \citet{Silver2017, Silver2018}.  
Furthermore, a measure with a loading on variance (defined in Section \ref{sec:method}) is related to UCB-V algorithm developed in \citet{audibert2007tuning}. 
Another set of methods has been developed in an attempt to solve notoriously hard Montezuma's Revenge, see for example \citet{GoExplore19, moczulskimontezuma2019}.

In our work we also use hindsight, see \citet{hind19}. Its primarily motivation is to enrich the learning signal and train a universal value function. However, from a certain point of view it can be seen as an exploration algorithm.

\section{Method}
\label{sec:method}
In this section, we describe our method: the planning and exploration components, as well as the training protocol. 
Algorithm \ref{alg1} shows how the components are brought together in the training loop.
The pseudo-code for $\mathtt{planner.run\_episode}$ is listed in Algorithm \ref{alg:run_episode} and, for the sake of clarity, the remainder of pseudo-code was moved to Appendix~\ref{sec:Algorithm}. For the model-based planer component, we develop an MCTS-inspired algorithm.\footnote{
MCTS is a family of model-based algorithms that iteratively build a search tree, alternating between the following stages: tree traversal, leaf expansion and evaluation, and backpropagation, see \citet{browne2012survey}
for a survey on the topic.}

The main novelty is using a risk-sensitive policy (for tree traversal and action choice) intended to guide exploration. We also investigate techniques exploiting the graph structure, including cycle avoidance \footnote{This is useful for a broad class of environments (including the ones we used in the experimental part) for which the optimal trajectory does not have cycles.} and develop a novel method of calculating targets for value function training. 


\begin{algorithm*}
    \caption{Learning and planning with ensembles}
    \label{alg1}
\begin{algorithmic}[1]
    \Require Environment $\mathtt{env}$, Model $\mathtt{model}$
    \State Initialize parameters of value function ensemble $\theta=(\theta_1, \ldots, \theta_K)$, $\mathbf{V}_\theta=(V_{\theta_1}, \ldots,V_{\theta_K})$    
    \State Initialize $\mathtt{replay\_buffer}$
    \Repeat
        \State $\mathtt{s \gets env.reset()}$
        \State $\mathtt{episode, solved}\gets \mathtt{planner.run\_episode(s; model,\mathbf{V}_\theta)}$ \Comment{see Algorithm \ref{alg:run_episode}}
        \State $\mathtt{values}\gets \mathtt{evaluate\_episode(episode)}$  \Comment{see Algorithm \ref{alg:evaluate_episode}}
        \State Optionally calculate a $\mathtt{mask}$ \Comment{see Appendix \ref{sec:training_details}} 
        \State $\mathtt{replay\_buffer.add(episode, values, solved, mask)}$ \Comment{see Algorithm \ref{alg:replay_buffer_add}} 
                \State $B\gets \mathtt{replay\_buffer.batch()}$
                \Comment{$B=\{(s_b, v_b, m_b)\}$, see Algorithm \ref{alg:replay_buffer_batch}}
                \State Update $\mathbf{V}_\theta$ by one step of gradient descent \Comment{e.g. RMSProp}
                \begin{equation*}
                    \nabla_{\theta_i}\left( \frac{1}{|B|}\sum_{(s,v,m) \in B} m_i\big( V_{\theta_i}(s) - v \big)^2+\zeta||\theta_i||^2\right), \qquad\text{for }\qquad i\in\{1, \ldots, K\}
                \end{equation*}
    \Until{convergence}
\end{algorithmic}
\end{algorithm*}

Loop avoidance is an extension of the transposition table techniques, see \citet{Childs2008, Gelly2012, Swiechowski2018} and is closely related to the virtual loss method of \citet{Segal11, McAleer19}. It is achieved in two ways: by backpropagation of some fixed negative value through the in-tree path ending with a leaf having no unvisited neighbors (see the pseudo-code in Appendix \ref{sec:Algorithm} related to $\mathtt{dead\_ends})$, 
and during tree traversal, when the agent is encouraged to avoid actions leading to previously visited states on the path (see  pseudo-code related to $\mathtt{penalty}_p$ in Appendix \ref{sec:Algorithm}). To strengthen this effect, the agent is also encouraged to avoid actions leading to previously visited states on the episode level (see parameter $\mathtt{penalty}_e$ in Algorithm \ref{alg:run_episode}).

These enhancements combined make it possible to learn even in sparse rewards scenarios, which is experimentally demonstrated in Section \ref{sec:experiments} (all used environments have sparse rewards). We note that Algorithm \ref{alg1} is not MCTS specific and other planners could be used as well.

The logic of our MCTS is laid out in Algorithm \ref{alg:run_episode}. We assume that any node of the search tree, say $\mathtt{n}$, stores a visit count $\mathtt{n.count}$, accumulated value $\mathtt{n.value}$, accumulated reward $\mathtt{n.reward}(a)$, and its children are denoted by $\mathtt{n.child}(a)$ for each action $a\in\mathcal A$.
We define 
\[
\widehat{Q}_\theta(\mathtt{n},a)=\mathtt{n.reward}(a)+\gamma\mathtt{n.child}(a)\mathtt{.value},\]
where $\gamma > 0$ is a discount rate. 

Our proposed risk-sensitive exploration method is implemented in Algorithm \ref{alg:choose_action}, see line $7$. It is used by the planner both in the action selection step (Algorithm \ref{alg:run_episode}, line $9$) and during the tree traversal stage (Algorithm \ref{alg:tree_traversal}, line $5$).
Its key defining elements are uncertainty modeling and risk measurement. A risk-sensitive tree traversal policy is defined as follows:
\begin{align}\label{eq:action_choose}
	a^*(\mathtt{n}) & :=\argmax_a \mathbb E_{\theta\sim \Theta}\left[\phi_a(\widehat{\mathbf{Q}}_\theta(\mathtt{n}))\right]\\
	\widehat{\mathbf{Q}}_\theta(\mathtt{n}) & := \left(\widehat{Q}_\theta(\mathtt{n},a')\colon a'\in\mathcal A\right),\nonumber
\end{align}
where $\mathcal{A}$ is the action space, $\phi_a:\mathbb R^{|\mathcal A|}\to \mathbb R$ is a risk measure, and $\widehat{Q}_\theta$ is an estimator of the $Q$-function. The posterior distribution $\Theta$ models uncertainty in the value estimation. 

We approximate 
the expected value from equation \ref{eq:action_choose}, 
$\mathbb E_{\theta\sim \Theta}\left[\phi_a(\widehat{\mathbf{Q}}_\theta(\mathtt{n}))\right]$, using Monte-Carlo and ensembles, by 
$\frac{1}{K}\sum_{i=1}^K \phi_a(\widehat{\mathbf{Q}}_{\theta_i}(\mathtt{n}))$, where $K$ is the size of ensemble. This approximation is motivated by \citet[Lemma 3]{Osband2018a}, see Appendix \ref{sec:training_details}. In our case we use an ensemble of value functions. In some cases, we "sub-sample" from $\Theta$, which is inspired by \cite{OsbandBPR16} and the classical Thomson sampling (see Appendix \ref{sec:training_details} for details).

\noindent
\begin{minipage}[t]{.46\textwidth}
\begin{algorithm}[H]
\caption{$\mathtt{planner.run\_episode()}$}
\label{alg:run_episode}
\begin{tabular}{ll}
\textbf{Require:} 
& $\mathtt{max\_episode\_len}$  \\
& $\mathtt{num\_mcts\_passes}$  \\
& $\mathtt{model}$ \\
& $\mathtt{penalty}_e$ \Comment{Episode penalty} \\
& $\mathbf{V}_\theta$  \Comment{Value function}\\ 
\textbf{Input:} &$\mathtt{s}$ \Comment{Starting state}\\ 
\end{tabular}
\begin{algorithmic}[1]
    \State $\mathtt{episode}\gets \emptyset$
    \For{$\mathtt{step}=1\text{ to }\mathtt{max\_episode\_len}$}
        \State $\mathtt{root}\gets \mathtt{s}$
        \State $\mathtt{root.value\gets root.value-}\mathtt{penalty}_e$ 
        \For{$1\text{ to }\mathtt{num\_mcts\_passes}$}  
            \State $\mathtt{path,leaf\gets traversal(root)}$
            \State $\mathtt{value\gets expand\_leaf(leaf;\mathbf{V}_\theta;model)}$ 
            \State $\mathtt{backpropagate(value,path)}$
        \EndFor
        \State $\mathtt{a\gets choose\_action(root,\{root\})}$
        \State $\mathtt{s, r, done\gets env.step(a)}$
        \State $\mathtt{episode.append((root, a, r))}$
        \If{$\mathtt{done}$} \textbf{break}
        \EndIf
    \EndFor
    \State \Return $\mathtt{episode, done}$ 
\end{algorithmic}
\end{algorithm}
%
\end{minipage}
\hfill
\begin{minipage}[t]{.51\textwidth}
\begin{algorithm}[H]
    \caption{$\mathtt{choose\_action()}$}\label{alg:choose_action}
    \textbf{Require:} $\mathtt{avoid\_loops}$\Comment{Bool}\\
    \textbf{Input:} $\mathtt{n}, \mathtt{seen}$
\begin{algorithmic}[1]
    \If{$\mathtt{avoid\_loops}$}
        \State $A\gets\{a\in\mathcal A(\mathtt{n})\colon\mathtt{n.child(a)\notin seen}\}$
        \If{$A = \emptyset$} \Comment{Terminal or dead-end}
            \State \Return $\mathtt{None}$
        \EndIf
    \Else
        \State $A\gets \mathcal A(\mathtt{n})$
    \EndIf
    \State Choose action according to \eqref{eq:action_choose} from set of available actions $A$
    \[
    a^*\gets \argmax_{a\in A} \mathbb E_{\theta\sim \Theta}\left[\phi_a(\widehat{\mathbf{Q}}_\theta(\texttt{n}))\right]
    \]
    \State \Return $a^*$
\end{algorithmic}
\end{algorithm}
\end{minipage}
\medskip

Utilizing a risk measure is inspired by \citet{Lowrey2018}, who used $\phi_a(x)=e^{\kappa x_a}$ for $x\in \mathbb R^{|\mathcal A|}$ and $\kappa>0$. In this paper we consider the following choices of $\phi_a$:
\begin{itemize}
    \item A measure with a loading on variance, $\phi_a(x)=x_a + \kappa x_a^2$, $\kappa>0$. This includes second moments and can be easily generalized to include variance, standard deviation and exploration bonuses.
    \item A relative majority vote (also known as plurality vote) measure,  
    \begin{equation}\label{eq:voting}
        \phi_a(x)= 1\left(\argmax_{a'}x_{a'}=a\right).
    \end{equation}
     Contrary to the other cases, $\phi_a$ defined in \eqref{eq:voting}, depends not only on marginal values of its input, but the whole input (i.e. the estimator vector $\widehat{\mathbf{Q}}_\theta$). It leads to a rule resembling optimal Bayes classifier form, i.e. the one which chooses $a$ minimizing $\mathbb P(a^*_\theta(s)\ne a)$.  
\end{itemize}

The intuitions behind the aforementioned choices of $\phi_a$'s are as follows. 
One can think of $\phi_a(x) = e^{\kappa x_a}$, used in \citet{Lowrey2018}, as a measure capturing all moments of value ensemble. For small values of $\kappa\approx 0$, it behaves like the measure with a loading on variance (via the Tylor expansion). The mean approximates the  value of a given action, while the variance quantifies the epistemic uncertainty. Taking weighted sum of these terms has the aim of balancing exploitation and exploration.
It is also related to UCB-V algorithm, see \citet{audibert2007tuning}. 
Voting, on the other hand, is a well established approach when combining ensembles, see e.g. \citet{breiman1996bagging} or \citet{rokach2010ensemble}.
A relative majority vote, in particular, is simple and it can lead to good performance, see e.g. in \citet[Section 6.4]{OsbandBPR16}. In the context of planning and RL,  it has several interesting properties. In particular, the distribution of votes across ensembles encodes the uncertainty related to the optimal action in a given state. 
High uncertainty may result in stochastic movement (caused by uniform tie breaking), and consequently lead to higher exploration. 
On the other hand, low uncertainty may result in an exploitative behavior of an agent and, as a result, less exploration. Additionally, voting may improve decision making of an agent in the states, where some action can be dangerous (e.g. lead to irreversible states).

We conclude this section by describing the part of Algorithm \ref{alg1} related to the update of value functions. The episodes are collected and stored in a prioritized replay buffer. The replay buffer performs some bookkeeping by storing, for each transition, information whether it comes from a solved episode or not (in the case of environments that provide such information, like Sokoban). This information can be used to prioritize experience: we use it to sample batches with a fixed ratio of solved to unsolved transitions 
(see Algorithm \ref{alg:evaluate_episode} in Appendix \ref{sec:Algorithm}).
Such a method resembles self-imitation techniques, see e.g. \citet{sil2018}.

The value functions are trained to minimize $l_2$ distance from target values sampled from the buffer. 
The target values can be computed in two modes: $\mathtt{bootstrap}$, which utilizes the values accumulated during the MCTS phase, and $\mathtt{factual}$, corresponding to discounted rewards in an episode (see Algorithm \ref{alg:evaluate_episode} in Appendix \ref{sec:Algorithm}). Up to our knowledge, the mode $\mathtt{bootstrap}$ is new. It generated a noticeable improvement of training performance becoming  the default choice in our main experiments (see Appendix \ref{sec:valueTargetAblation} for comparison).

Some experiments use masks which, analogously to \cite{Osband2018a}, form a mechanism of assigning a transition to a value function (see Appendix \ref{sec:training_details}).

\section{Experiments}\label{sec:experiments}

In this section, we provide experimental evidence to show that using ensembles and risk measures is useful. We chose three environments: Deep-sea, toy Montezuma's Revenge, and Sokoban. In all cases we work with sparse reward versions of the environments i.e. the agent's is rewarded only upon successful completion of the task. 

We use an MCTS planner with the number of passes equal to $10$ (see line $4$ of Algorithm \ref{alg:run_episode}), which is rather modest for MCTS-like planning methods.\footnote{For example, the recent work \citet{Schrittwieser2019} uses $800$ passes to plan in board games.} Interestingly,  we observed that such a relatively weak planner is sufficient to obtain a well-preforming algorithm. 

In the Sokoban multi-board experiments, we use a learned model; otherwise we assume access to the perfect model. A good model for the Sokoban environment could be acquired from random trajectories, which is improbable for other domains. Extending our algorithm to these cases is an exciting research direction, which will require development of planning methods robust to model errors, possibly again using ensembles (see e.g. \citet{Kurutach2018}) and training the model and the policy in an interlaced fashion (e.g. similar to \citet{Kaiser19}).

We utilize various neural network architectures, see Appendix \ref{sec:architectures}. We measure uncertainty using standard deviation except for the case of Sokoban with randomly-generated boards, where voting was used, see \eqref{eq:voting}. Configuration of the experiments is summarized in Table \ref{tab:hiperparameters} in Appendix \ref{sec:training_details}. 

\subsection{Deep-sea}

Deep-sea environment was introduced in \citet[Section 4]{Osband2018a} and later included in \citet{bsuit} as a benchmark for exploration. 
The agent is located in the upper-left corner (position $(0,0)$) on a $N\times N$ grid, $N\in \mathbb{N}$. 

\def\pmsize{0.22}
\def\pmsizetwo{0.95}
\begin{figure*}[t]
  \centering
\begin{subfigure}{\pmsize\textwidth}
  \centering
  \includegraphics[width=\pmsizetwo\linewidth]{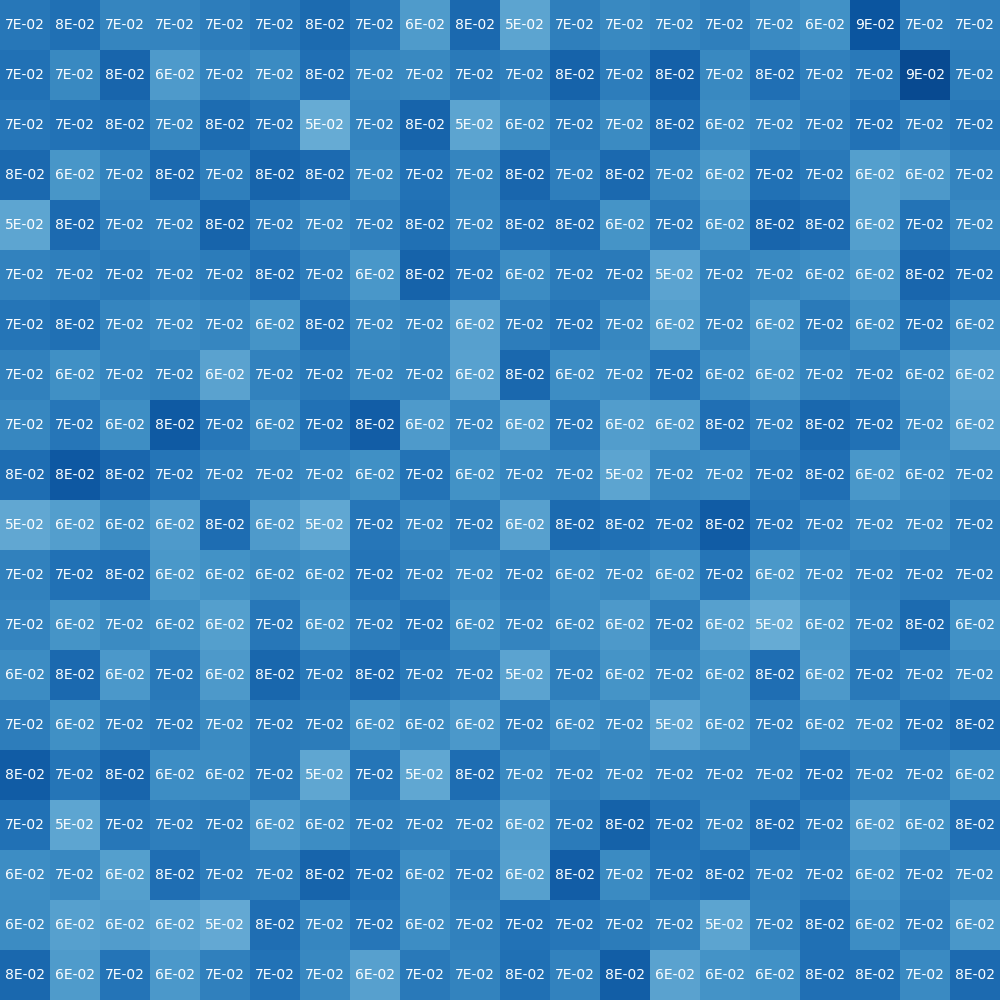}
  \caption*{step $0$}
\end{subfigure}
\begin{subfigure}{\pmsize\textwidth}
  \centering
  \includegraphics[width=\pmsizetwo\linewidth]{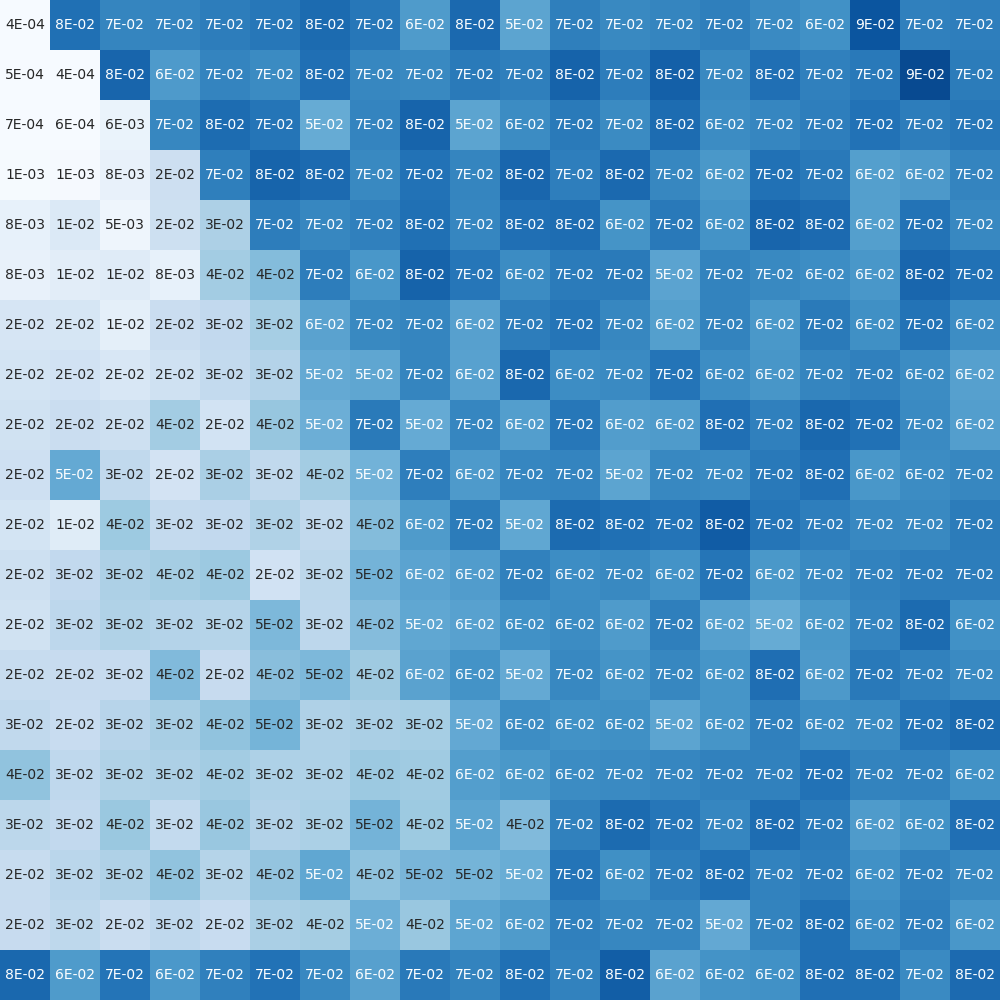}
  \caption*{step $400000$}
\end{subfigure}
\begin{subfigure}{\pmsize\textwidth}
  \centering
  \includegraphics[width=\pmsizetwo\linewidth]{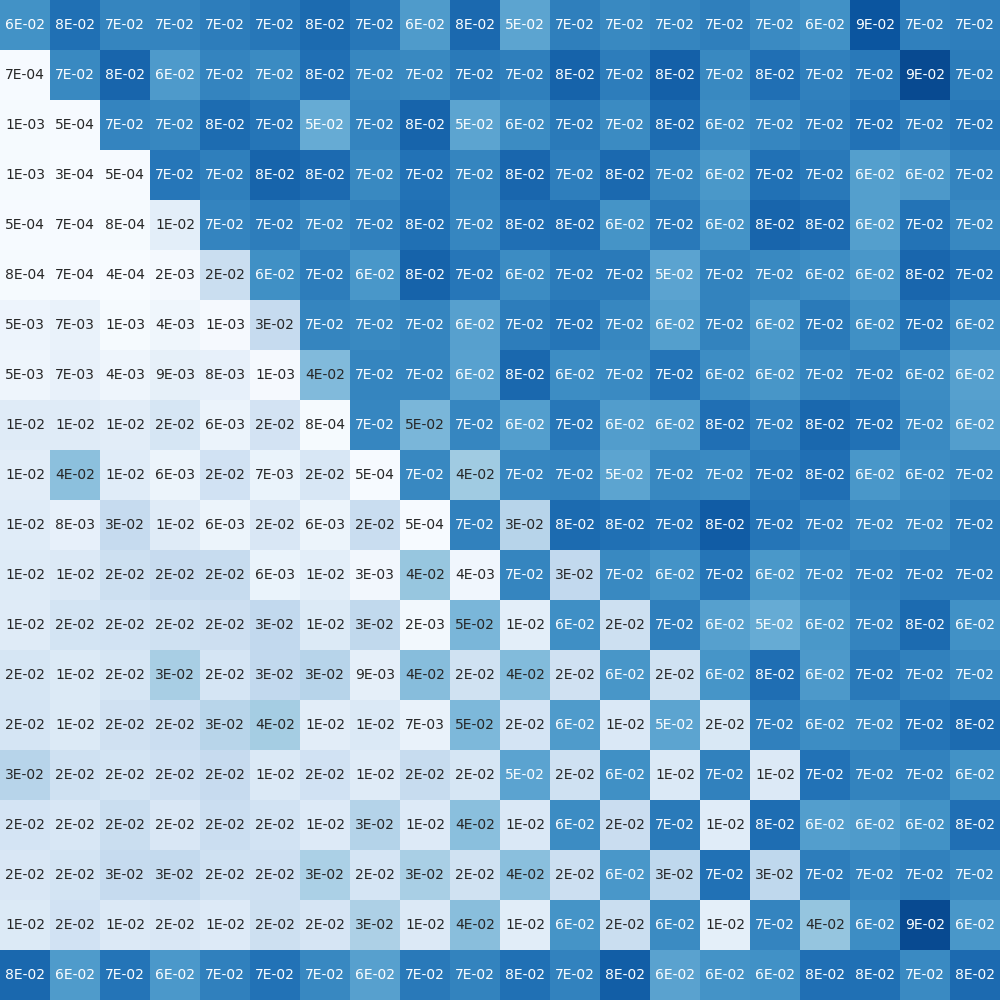}
  \caption*{step $1200000$}
\end{subfigure}
\begin{subfigure}{\pmsize\textwidth}
  \centering
  \includegraphics[width=\pmsizetwo\linewidth]{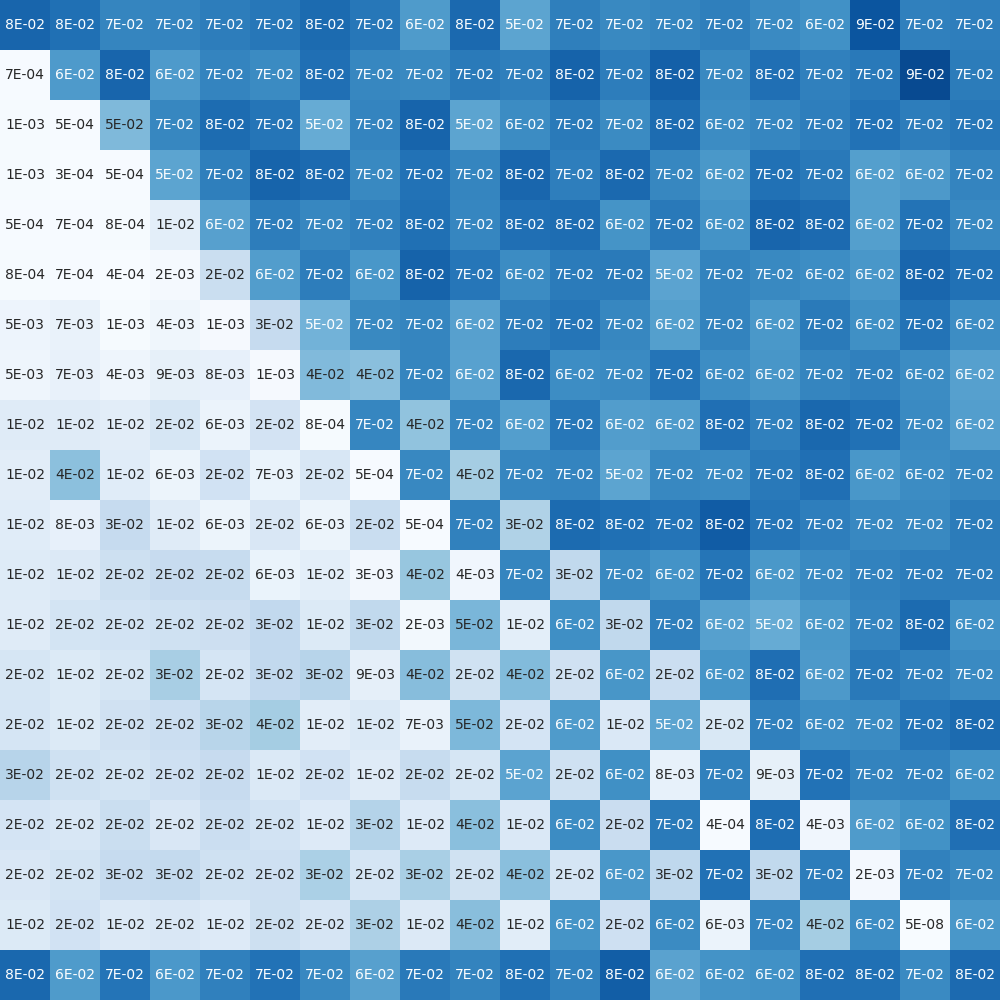}
  \caption*{step $1600000$}
\end{subfigure}
\caption{The heatmaps of standard deviations of ensemble values in the Deep-sea environment. High values are marked in blue and low in white. At the beginning of training (left picture) the standard deviation is high for all states. Gradually it is decreased in the states that have been explored. Finally (the right) the reward state is found. Note that the upper-right part of the board is unreachable.}\label{fig:heatmaps}
\end{figure*}

In each timestep, its $y$-coordinate is increased, while $x$ is controlled. The agent issues actions in $\{-1, 1\}$. These are translated to \emph{step left} or  \emph{step right} 
(increasing $x$ by $-1$ or $+1$, respectively, as long as $x\ge0$; otherwise $x$ remains unchanged) according to a prescribed action mask (not to be confused with transition masks in Algorithm \ref{alg1}). 
For each \emph{step right} the agent is punished with $0.01/N$. After $N$ steps, the game ends, and the agent receives reward $+1$ if and only if it reaches position $(N, N)$. The action mask mentioned above is randomized at each field at the beginning of training (and kept fixed). 

\begin{figure*}
	\includegraphics[width=0.95\textwidth]{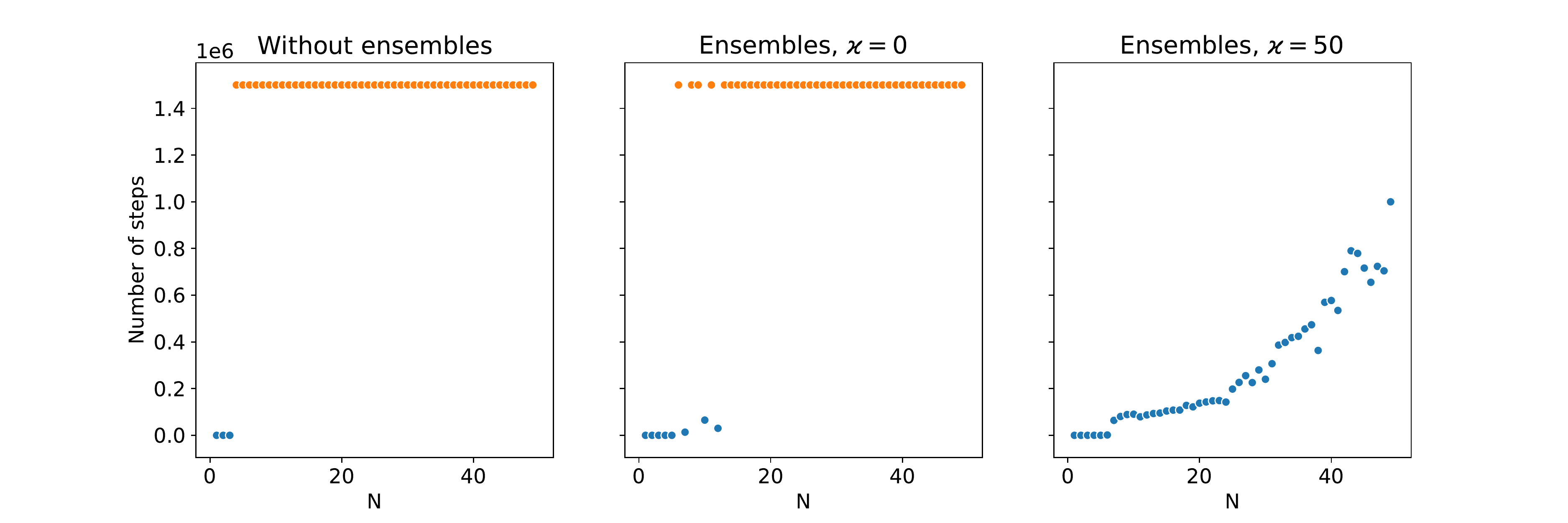}
	\caption{Comparison of number of steps needed to solve the deep-sea environment with given grid size $N$. Orange dots marks trials which were unable to solve problem in 1400000 steps. Large problem instances ($N>20$) are solved only when exploration bonus is used (right-most plot, $\kappa=50$). }\label{fig:deepsea}
	\vspace{-15pt}
\end{figure*}
Such a game is purposely constructed so that naive random exploration schemes fail already for small $N$'s. Indeed, a random agent has chance $(1/2)^N$ of reaching the goal even if we disregard misleading rewards for \emph{step right}. 
The exploration progress for our method is shown in 
Figure~\ref{fig:heatmaps}. In Figure~\ref{fig:deepsea}, one can see a comparison of non-ensemble models, ensemble models with Thomson sampling (see Appendix \ref{sec:training_details}) but without uncertainty bonus ($\kappa=0$), and our final ensemble model with uncertainty bonus $\kappa=50$. We conclude that using both sub-sampling and ensembles is essential to achieving good exploration (for details see also Appendix \ref{sec:training_details}).

\subsection{Toy Montezuma’s Revenge}\label{subsec:toymr}
Toy Montezuma's Revenge is a navigation maze-like environment. It was introduced in \citet{Roderick2018} to evaluate ideas of using higher-level abstractions in long-horizon problems. 

\begin{figure}[h!]
  \centering
    \includegraphics[width=0.4\columnwidth]{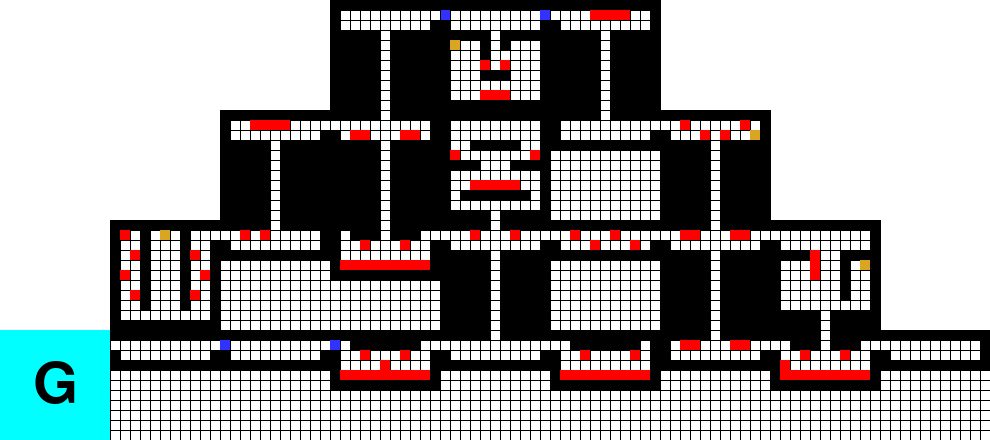}
    \caption{The biggest toy Montezuma's Revenge map, consisting of $24$ room. The goal is to reach the room marked with $G$. The agent needs to avoid traps (marked in red) and pass through doors (marked in blue). Doors are open using keys (marked in yellow). 
    }\label{fig:toy-mr}
\end{figure}

While its visual layer is greatly reduced version of the actual Montezuma's Revenge Atari game, it retains much of the original's exploration problems.   This makes it a useful test environment for exploration algorithms, see e.g. \citet{moczulskimontezuma2019}. In our experiments we work with with the biggest map with $24$ rooms, see Figure \ref{fig:toy-mr}.\footnote{We use a slightly modified code from \url{https://github.com/chrisgrimm/deep_abstract_q_network}} In order to concentrate on the evaluation of exploration we chose to work with sparse rewards. The agent gets reward $1$ only if it reaches the treasure room, otherwise the episode is terminated after $300$ steps.

It is expected that any simple exploration technique would fail in this case (we provide some baselines in Table \ref{table:montezuma-results}). \citet{moczulskimontezuma2019} benchmarks PPO, PPO with self imitation learning (PPO+SIL), PPO with count based exploration bonus (PPP+EXP) and their new technique (DTSIL). Only DTSIL is consistently able to solve $24$ room challenge, with PPO+EXP occasionally reaching this goal.  Our method based on ensembles and model-based planning solves this exploration challenge even in a harder, sparse reward case.\footnote{DTSIL builds on the intermediate partial solutions, which are ranked according to their reward, thus we suspect it would fail in the sparse reward case.} The results are summarized in Table \ref{table:montezuma-results}. We have three setups: \emph{no-ensemble}; \emph{ensemble, no std}; \emph{ensemble, std}. 
In the first case, we train using Algorithm \ref{alg1} with a single  neural-network. In the second case, for each episode we sub-sample $10$ members of an ensemble of size $20$ to be used and MCTS is guided by their mean. In the final, third case, we follow the same protocol but we add to the mean the standard deviation. 
In our experiments we observe that \emph{no-ensemble} in $30$ out of $43$ cases does not leave behind the first room. The setup without explicit exploration bonus, \emph{ensemble, no std}, perform only slightly better. Finally, we observe that \emph{ensemble, std} behaves very well.

Further experimental details and the network architecture are presented in Appendix \ref{sec:architectures} and \ref{sec:training_details}. 

\begin{table}
\begin{tabular}{ccc}\\\toprule  
Setup & Solved / no. seeds & Av. visited rooms \\\midrule
\emph{no-ensemble} & 0 / 43 & 4.7 \\  
\emph{ensemble, no std} & 2 / 40 & 5.8 \\  
\emph{ensemble, std} & 30 / 37  & 17.5\\  \bottomrule
\end{tabular}
\caption{Result for toy Montezuma's Revenge. We report the number of runs which found solution in $1.2e6$ steps and the number of seeds of network initialization. We also show the average number of visited rooms, which is a proxy of the learning progress. }\label{table:montezuma-results}
  \vspace{-20pt}
\end{table}

\subsection{Sokoban}\label{sec:sokoban}
\begin{wrapfigure}{R}{0.3\textwidth} 
 \vspace{-15pt}
  \begin{center}
    \includegraphics[width=0.3\textwidth]{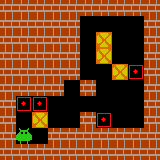}
    \caption{Example $(10, 10)$ Sokoban board with $4$ boxes. Boxes (yellow) are to be pushed by agent (green) to designed spots (red). The optimal solution in this level has $37$ steps.}\label{fig:sokoban}
    \label{fig:databaseUserTable}
  \end{center}
  \vspace{-15pt}
\end{wrapfigure}

Sokoban is an environment known for its combinatorial complexity. The agent's goal is to push all boxes (marked as yellow, crossed squares) to the designed spots (marked as  squares with a red dot in the middle), see Figure \ref{fig:sokoban}. Apart from the navigational challenge, the difficulty of this game is greatly increased by the fact that some actions are irreversible. A canonical example of such an action is pushing a box into a corner, though there are multiple less obvious cases. Formally, this  difficulty manifests itself in the fact that deciding whether a level of Sokoban is solvable or not, is NP-hard, see e.g. \citet{dor1999sokoban}. Due to these challenges the game has been considered as a testbed for reinforcement learning and planning methods. 

Operationally, to generate Sokoban levels we use an automated procedure
proposed by \citet{RacaniereWRBGRB17}. Some RL approaches to solving Sokoban (e.g. \citet{RacaniereWRBGRB17, Guez19}) assume additional reward signal in the game, e.g. for pushing a box into a designed spot. In this work we use sparse setting, that is  the agent is rewarded only when all the boxes are put in place.

It is interesting to note, that Sokoban offers two exploration problems:
single-level-centric, where a level-specific exploration is needed, and multi-level-centric, where a 'meta-exploration' strategy is required, which works in a  level-agnostic manner or can quickly adapt. 
As a result, we conducted three lines of experiments using ensembles: \textit{ a) learning to solve randomly generated boards (dubbed as multiple-boards Sokoban), 
b) learning to solve a single board (dubbed as single-boards Sokoban) , c) transfer and learnable ensembles. }

In our experiment we use Sokoban with board of size $(10, 10)$ and 4 boxes. We use the limit of $200$ steps in the experiment \emph{a)} and $100$ in the remaining ones.

\paragraph{Multiple-board Sokoban: learning to solve randomly generated boards} 
\begin{figure}
  \begin{center}
    \includegraphics[width=1.\columnwidth]{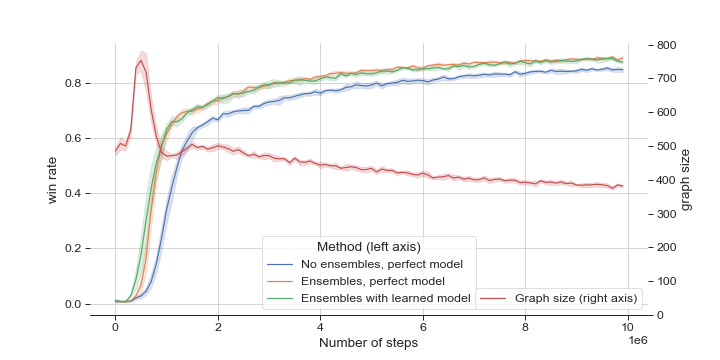}
    \caption{Learning curve (left axis) and the size of explored graph (right axis) of Sokoban states. 
    The shape of the latter plot may show a gradual switch from exploration to exploitation. The results are averaged over 10 random seeds (5 for experiments with learned model), shaded areas shows 95\% confidence intervals. 
    }\label{fig:sok-ens}
  \end{center}
  \vspace{-15pt}
\end{figure}
In this experiment we measure the ability of our approach to solve randomly generated Sokoban boards. We show also that the approach is flexible enough to accommodate for the use of a learned model of dynamics. More precisely, the planning described in Algorithm \ref{alg:run_episode} is performed using this model.

The model is trained using a set of trajectories obtained by a random (uniform) agent. The major difficulty in obtaining the model is learning (sparse) rewards. The ratio of non-zero reward transitions is less than 3e-6. To tackle this problem we generated a dataset consisting of boards with $1$, $2$, $3$, $4$ boxes and substantially upsample rewarded transitions. Details are provided in Appendix \ref{sec:training_details}. 

To measure uncertainty we use an ensemble value function using relative majority voting as formalized in \eqref{eq:voting}. Relative majority voting takes into account the uncertainty of ensembles when it comes to the final outcome, not only the uncertainty in assessment of particular action.

After $10$ million steps in the real environment, our method reaches $89.0$\% win rate, compared to $84.7$\% of an agent not using ensembles, see Figure~\ref{fig:sok-ens} (the win rate is calculate on the last $1000$ games). We also see that using a learned model yields result practically equivalent to that of the perfect model.

As a measure of exploration we also present the size of the game graph explored during episodes (red curve in Figure \ref{fig:sok-ens}). It shows an interesting effect, which can be interpreted as a transition from exploration to exploitation approximately at a $40\%$ win-rate mark or, equivalently, $10000$ games.

\paragraph{Single-board Sokoban: learning to solve a single board} In this experiment we measure the extent in which our methods can plan and learn on single boards. We note that this setting differs substantially from the one in the previous paragraph. In the multiple-boards Sokoban there is a possibility of generalization from easier to harder scenarios (e.g. MCTS with randomly initialized value network solves $\approx 0.7\%$ of boards). Such a transfer is not possible in the single-board case studied here. On the other hand, the algorithm gathers multiple episodes from the same board, which enables the agent to explore the board's state space.

For the single-board Sokoban we used experimental settings similar to the one in Section \ref{subsec:toymr}, see also details in Appendix \ref{sec:training_details}. We observe that the setup with ensembles solves $73\%$ compared to $50\%$ the standard training without ensembles (experiments with and without ensembles were performed on a common set of $250$ randomly generated boards). The latter might seem surprisingly good, taking into account the sparse reward. This follows by the loop avoidance described in Section \ref{sec:method}. If during planning the agent finds itself in a situation from which it cannot find a novel state (i.e. encounters $\mathtt{dead\_end}$ in Algorithm \ref{alg:expand_leaf}) a negative value, set to $-2$ in our experiments, is backpropagated in Algorithm \ref{alg:backpropagate} to already seen vertices. We speculate that this introduces a form of implicit exploration. 

In the singe-board Sokoban case we performed also experiments on $(8,8)$ boards with $4$ boxes obtaining success rate of $94\%$ compared to $64\%$ on the standard non-ensemble settings.

\paragraph{Sokoban: transfer and learnable ensembles} Generating any new board can be seen as a cost dimension along with sample complexity. This quite naturally happens in meta-learning problems. 
We tested how value functions learned on small number of boards perform on new, previously unseen, ones. We used the following protocol: we trained value function on fixed number of $10$ games. To ease the training, we used relabelling akin to \citet{hind19}\footnote{More precisely, for a failing trajectory we choose a random time-step and shift the target spots so that they match the current location of boxes. We note that although this operation requires the knowledge of the game mechanics (i.e. its perfect model) it is used only in this phase}. 
\begin{table}
	  \begin{tabular}{ccccc}
		  \toprule
	      Architecture & Random & Trans. $1$ & Trans. $2$ & Trans. $3$ \\ \midrule
		 $5$-layers &  $0.7\%$ & $4.9\%$  & $7.1\%$ & $8.5$ $\%$\\
		 $4$-layers &  $0.7\%$ & $4.3\%$  & $5.6\%$  & $7.3\%$ \\
		 \bottomrule
	  \end{tabular}
	  \caption{Results of transfer experiments. We test transfers from one value function (Trans. $1$) and transfer from ensembles of $2$ and $3$ value functions (Trans. $2$ and Trans. $3$). In the later two cases the aggregation of values is learned. The results are averaged over $20$ seeds.}\label{table:transfer_results}
  \vspace{-10pt}
\end{table}
We evaluated these functions on other boards. It turns out that they are typically quite weak, which is not very surprising, as solving $10$ boards does not give much chance to infer 'the general' solutions. Next, we used ensembles of the value functions. 
More precisely, we calculated the values of $n=2,3$ models and aggregated them using a small neural networks with one fully connected hidden layer. This network is learnable and trained using the standard setup. We observe that the quality increases with the number of value functions in the ensemble as summarized in Table \ref{table:transfer_results}. We observed high variability of the results over seeds, which is to be expected as Sokoban levels significantly vary in difficulty. We also observe that maximal results for transfer increase with the number of value function, being approximately $10\%$, $11\%$ and $12\%$. This further supports the claim that ensembling may lead to improved performance. In $5$-layer experiment we use a network with $5$ hidden convolutional neural network layers, see details in Appendix \ref{sec:architectures}, we compare this with an analogous $4$ layers network. In the latter case, we obtain a weaker result. We speculate this might be due to the fact that larger networks generalize better, see e.g. \citet{Cobbe19}.

\section{Conclusions and further work}\label{sec:conclusions}

In this paper, we introduced a reinforcement learning method that blends planning, learning, and risk-sensitive approach to exploration. We verified experimentally that such a setup is useful 
in solving hard exploration problems, i.e. problems characterized by sparse rewards and long episodes (e.g. spanning even hundreds of steps).

We believe that this opens promising future research directions. There are multiple ensemble design choices, and we tested only a selected few. Additionally, there are more ways to combine the results of ensembles and it would be interesting to 
see if one, relatively general, method can be found. Such a result would be a step towards deep Bayesian learning. 

In our work, we used a learned model in the case Sokoban, in which a relatively good model could be obtained from random trajectories. It would to interesting to cover a general case, when learning model and agent's behaviour needs to occur simultaneously. Perhaps the most challenging problem is making planner robust to model errors. Equally important research direction would be related to solving stochastic environments. This might be considerably more difficult as such a task requires disentangling epistemic (studied in this work) and aleatoric (coming from the environment) uncertainties. 

We focused our attention on MCTS, but there is a priori no reason why some other planning method should not yield better results. In some initial experiments we obtained promising, but yet inconclusive, results using the Levin tree search (see \citet{Orseau2018}). Another tempting direction is training both value function and a policy, akin to methods of \citet{Silver2017}.

It would also be interesting to isolate the proposed exploration method from the planner and see how it fares when coupled with model-free algorithms. This fits along the lines of some recent research directions, see e.g.  \citet{agarwal2019striving}.

Going further, we speculate that it may be possible to use the methods developed for Sokoban in meta-learning and continual learning grounds problems, perhaps akin to recent \citet{Lu2019}. Measures of uncertainty can possibly enable a learning system to adapt to a changing environment. In an archetypical case, this might be obtained by choosing from ensemble a model (a skill) which is useful at the moment and understanding situations that such a model is not yet present.

\subsection*{Acknowledgments} This research was supported by the PL-Grid Infrastructure. We extensively used the Prometheus
supercomputer, located in the Academic Computer Center Cyfronet in the AGH University of Science and Technology in Kraków, Poland. The work of Konrad Czechowski, Piotr Kozakowski and Piotr Miłoś was supported by the Polish National Science Center grants UMO-2017/26/E/ST6/00622. We managed our experiments using the Neptune tool \url{https://neptune.ai/}. We would like to thank the Neptune team for providing us access to the team version and technical support

\bibliography{bibliography}
\bibliographystyle{icml2019}

\newpage
\onecolumn
\setcounter{section}{0}
\renewcommand{\thesection}{A.\arabic{section}}
\section{Algorithm}
\label{sec:Algorithm}
In this section we detail the building blocks of Algorithm \ref{alg1}. Our algorithm takes into account the graph structure of the underlying environment. Following inspiration of the transposition table techniques, see \citet{Childs2008, Gelly2012, Swiechowski2018}, we calculate and store values corresponding to the nodes of the graph (i.e. the states of the environment). These nodes are clearly different than the nodes of the tree search. To ease the notation, we write $\mathtt{node.value}$, $\mathtt{node.value}$ instead of $\mathtt{node.state.value}$, $\mathtt{node.state.value}$, respectively (with the exception of Algorithm \ref{alg:expand_leaf}, where these are explicit). Importantly, in our approach these values are vectors of dimension equal to the size of ensemble.

As mentioned in Section \ref{sec:method}, the key elements of MCTS (see Algorithm \ref{alg:run_episode}) are: tree traversal, leaf expansion and backpropagation (shown in Algorithm \ref{alg:tree_traversal}, Algorithm \ref{alg:expand_leaf} and Algorithm \ref{alg:backpropagate}, respectively).

Algorithm \ref{alg:expand_leaf} shows how the model and value function enter the picture. The model is used for generating next states (line $10$) and the value function evaluates new, previously unvisited, states. The visited states are kept in a global transposition table. 
Algorithm \ref{alg:update} shows the update of a tree node.

Algorithm \ref{alg:tree_traversal} and Algorithm \ref{alg:backpropagate} both use variable $\mathtt{penalty}_p$.  This is a penalty corresponding to entering the same states during tree traversal stage (hence is operates on the planner level) and the change is applied during traversal, and undone during backpropagation. 

\noindent
\begin{minipage}[t]{.50\textwidth}
\begin{algorithm}[H]
    \caption{$\mathtt{traversal()}$} 
    \label{alg:tree_traversal}
    \textbf{Require:} $\mathtt{penalty}_p$ \Comment{Planner penalty}\\
    \textbf{Input:} $\mathtt{root}$
\begin{algorithmic}[1]
    \State $\mathtt{n\gets root}$ 
    \State $\mathtt{path\gets \emptyset}$
    \While{$\mathtt{n}$ is not a leaf} 
        \State $\mathtt{n.value\gets n.value-}\mathtt{penalty}_p$
        \State $a\gets \mathtt{choose\_action}(\mathtt{n}, \mathtt{path})$
        \If{$a$ is $\mathtt{None}$} \Comment{Dead-end}
            \State\textbf{break}
        \EndIf
        \State $\mathtt{path.append((n, }a))$
        \State $\mathtt{n \gets n.child(a)}$
    \EndWhile
    \State \Return $\mathtt{path}, \mathtt{n}$ \Comment{$\mathtt{n\notin path}$}
\end{algorithmic}
\end{algorithm}
\vspace{-1.5em}
\begin{algorithm}[H]
    \caption{$\mathtt{backpropagate()}$}
    \label{alg:backpropagate}
    \textbf{Require:} $\mathtt{penalty}_p$, $\gamma$\\
    \textbf{Input:} $\mathtt{v}$, $\mathtt{path}$
\begin{algorithmic}[1]
    \For{$(\mathtt{n}, a)$ in $\mathtt{reversed(path)}$}
        \State $\mathtt{n.value\gets n.value+}\mathtt{penalty}_p$
        \State $\mathtt{v\gets n.reward + \gamma v}$
        \State $\mathtt{update(n, v)}$
    \EndFor
\end{algorithmic}
\end{algorithm}
\end{minipage}
\hfill
\begin{minipage}[t]{.5\textwidth}
\begin{algorithm}[H]
    \caption{$\mathtt{expand\_leaf()}$}
    \label{alg:expand_leaf}
    \textbf{Require:} $\mathtt{dead\_end\_value}$, ${\mathbf{V}}_\theta$, $\mathtt{model}$ \\
    \textbf{Input:} $\mathtt{leaf}$\Comment{MCTS tree node without children}
\begin{algorithmic}[1]
    \If{$\mathtt{leaf}$ is terminal}
        \State $\mathtt{update(leaf, 0.)}$
        \State \Return 0.
    \ElsIf{$\mathtt{leaf}$ is a $\mathtt{dead\_end}$}
        \State $\mathtt{update(leaf, dead\_end\_value)}$
        \State \Return $\mathtt{dead\_end\_value}$
    \Else
        \For{$a\in\mathcal A(\mathtt{leaf})$}
			\State $\mathtt{new\_tree\_node} \gets \mathtt{create\_tree\_node}()$
            \State $\mathtt{next\_state} \gets \mathtt{model(leaf.state, a)}$
			\State  $\mathtt{new\_tree\_node.state} \gets \mathtt{next\_state}$
            \If{$\mathtt{next\_state}$ not yet visited}
                \State $\mathtt{next\_state.value}\gets {\mathbf{V}}_\theta(\mathtt{next\_state})$
            \EndIf
            \State $\mathtt{leaf.child(a)}\gets \mathtt{new\_tree\_node}$
        \EndFor
    \EndIf
    \State \Return $\mathtt{leaf.value}$
\end{algorithmic}
\end{algorithm}
\vspace{-1.5em}
\begin{algorithm}[H]
    \caption{$\mathtt{update()}$}
    \label{alg:update}
    \textbf{Input:} $\mathtt{n}$, $\mathtt{value}$
\begin{algorithmic}[1]
    \State $\mathtt{n.value}\gets \mathtt{n.value+value}$
    \State $\mathtt{n.count}\gets \mathtt{n.count}+1$
\end{algorithmic}
\end{algorithm}
\end{minipage}

The following blocks of code are related to the training setup. Algorithm \ref{alg:evaluate_episode} is responsible for computing an appropriate value for each element of the episode. 
There are two available modes: $\mathtt{bootstrap}$, which utilizes the values accumulated during the MCTS phase, and $\mathtt{factual}$, which represents the sum of discounted rewards in the episode. In the $\mathtt{bootstrap}$ mode we undo the penalty applied during the episode generation stage (line 5 in the Algorithm \ref{alg:run_episode}). We recall also that $s_t.value$ is a vector of dimension equal to the ensemble size; we take its mean in line 4. Up to our knowledge, the $\mathtt{bootstrap}$ mode is novel and performed favourably in experiments, see Appendix \ref{sec:valueTargetAblation}. 

The inner details of replay buffer, in particular Hindsight and prioritisation, are given in Algorithm \ref{alg:replay_buffer_add} and Algorithm \ref{alg:replay_buffer_batch}. Hindsight refers to any method that processes episode, and potentially overrides states values (see lines 1-2 in Algorithm \ref{alg:replay_buffer_add}). 
Algorithm \ref{alg:replay_buffer_batch} shows the way a batch is generated. 
First, for each transition it is determined whether it should be sampled from a solved or unsolved episode (according to a fixed ratio). 
Second, a game is sampled from a population determined in the previous step, with probability proportional to the game length. Finally, given a game a transition is chosen uniformly.

\noindent
\begin{minipage}[t]{.5\textwidth}
\begin{algorithm}[H]
    \caption{$\mathtt{evaluate\_episode()}$}
    \label{alg:evaluate_episode}
    \begin{tabular}{ll}
    \textbf{Require:} & $\mathtt{mode}$ \Comment{String}\\
     & $\mathtt{penalty}_e$\Comment{Episode penalty}\\
    \textbf{Input:} &$\mathtt{episode}$ \Comment $\{(s_t, a_t, r_t)\}$ \\
    & $\mathtt{solved}$\Comment Bool\\
    & $\gamma$ \Comment Discount rate
    \end{tabular}
    \begin{algorithmic}[1]
    \State $T\gets \mathtt{len(epsiode})$
    \State $\mathtt{values}\gets [0, \ldots, 0]\in\mathbb R^T$ 
    \If{$\mathtt{mode="bootstrap"}$}
    \State $\mathtt{values}\gets [\mathtt{mean}(s_t.\mathtt{value}), t\in {0,\ldots, T-1}]$  
    \State $\mathtt{values}\gets \mathtt{values} + \mathtt{penalty}_e$  
    \ElsIf{$\mathtt{mode="factual"}$}
    \State $\mathtt{r}\gets [0,\ldots,0,\mathbf{1}_\mathtt{solved}]$

    \For{$t=T-1\text{ to }1$} 
        \State $\mathtt{values}_{t-1}\gets \gamma \mathtt{values}_t+\mathtt{r}_t$ 
    \EndFor
    \EndIf
    \State \Return $\mathtt{values}$
\end{algorithmic}
\end{algorithm}
\end{minipage}%
\hfill
\begin{minipage}[t]{.47\textwidth}
\begin{algorithm}[H]
    \caption{$\mathtt{replay\_buffer.add()}$}
    \label{alg:replay_buffer_add}
    \begin{tabular}{ll}
    \textbf{Require:} & $\mathcal D$\Comment{Replay Buffer}\\ 
    & $H$\Comment{Hindsight mapping} \\
    \textbf{Input:} & $\mathtt{episode}$ \Comment{$\{(s_t, a_t, r_t, w_t)\}$}\\
    & $\mathtt{value}$ \Comment{$\{v_t\}$}\\
    & $\mathtt{solved}$ \Comment{Bool}\\
    & $\mathtt{mask}$ \Comment{Binary vector}
    \end{tabular}
\begin{algorithmic}[1]
    \For{$t=0\text{ to }T-1$}  \Comment{Hindsight}
        \State $v_{t}\gets H(\mathtt{episode})_t$ 
    \EndFor
    \State $\mathcal D\gets \mathcal D \cup (\{(s_t, a_t, v_t)\}, \mathtt{solved, mask})$ 
\end{algorithmic}
\end{algorithm}
\vspace{-1.5em}
\begin{algorithm}[H]
\caption{$\mathtt{replay\_buffer.batch()}$}
\label{alg:replay_buffer_batch}
\begin{tabular}{ll}
\textbf{Require:} & $\mathcal D$\Comment{Replay Buffer}\\ 
& $\mathtt{size}$\Comment{Batch size} \\
& $\mathtt{ratio}$\Comment{Solved/unsolved} \\
\end{tabular}
\begin{algorithmic}[1]
    \State Initialize $B\gets \emptyset$
    \For{$b=1\text{ to } \mathtt{size}$}
        \If{$b\times \mathtt{ratio}\%1=0$}
            \State $\mathtt{select\gets False}$
        \Else
            \State $\mathtt{select\gets True}$
        \EndIf
        \State $D'\gets \mathcal \mathcal \{d\in\mathcal D\colon d.\mathtt{solved=select}\}$
        \State Sample $d\in\mathcal D'$ with 
        \[
        \mathtt{prob}(d)\propto \mathtt{len}(d.\mathtt{episode}) 
        \]
        \State Sample $(s, v, m)$ uniformly from $d$
        \State $B\gets B\cup \{(s, v, m)\}$
    \EndFor
    \State \Return $B$ 
\end{algorithmic}
\end{algorithm}
\end{minipage}

\section{Architectures and input formats} \label{sec:architectures}

\paragraph{Deep-sea} For the Deep-sea environment we encode a state as a one hot vector of size $N^2$, and learn a simple linear transformation for the value estimation.

\paragraph{Toy Montezuma Revenge} Observations are represented as tuples containing the location of the current room, the agent's position within the room and the status of all keys and doors on the board. To estimate value, we use fully-connected neural networks with two hidden layers of $50$ neurons each.

\paragraph{Single-board Sokoban} Observations have shape $(10, 10, 7)$, where the first two coordinates are spatial and the third one is a one-hot encoding of the type of a state (e.g. box, target, agent, wall). To estimate the value, we flatten the observation and apply fully-connected neural networks with two hidden layers of $50$ neurones each.

\paragraph{Multiple-boards Sokoban} We use the same observation type as in the single-board problem. 
Each value function network is composed of five $3\times 3$ convolutional layers with with stride $1$, followed by two fully connected layers with $128$ units and $1$ unit, respectively. 

Now we describe details of the model architecture. The input consists of a one-hot representation of a board (as above) and an action in $\{1, \ldots, 4\}$. The one-hot representation of the action is concatenated with the board resulting in a tensor of shape $(10,10,11)$.  It is processed through two convolutional layers with the kernel sizes $(5, 5)$ and $64$ channels. We then apply a neural network with two heads. The first one applies a point-wise dense layer to produce the next Sokoban frame (again one-hot encoded). The second uses the global average pooling and a linear layer to predict if input state and action generates a reward (which in our sparse setup is equivalent to solving the board).

\paragraph{Sokoban: transfer and learnable ensembles} In the $5$-layer experiment we use the same architecture as in the multiple-boards Sokoban. In the $5$-layer we remove one of inner hidden convolutional layers. 

\section{Training details} \label{sec:training_details}

\paragraph{Masks} An important decision for training is how to assign transitions to the particular elements of value functions ensemble. This is implemented using masks, see Algorithm \ref{alg1}. 
Suppose a batch $B$ is to be used in an update step (see lines 9-10 of Algorithm \ref{alg1}) and let $\mathtt{t}=(s,v,m)\in B$ be a transition. 
A mask $m\in \{0,1\}^{\{1, \ldots, K\}}$ has  the following interpretation: $m_i = 1$ 
if and only if the transition $\mathtt{t}$ is used to train $V_{\theta_i}$, $i = 1,\ldots, K$. 

We experimented with the following versions of masking:
\begin{itemize}
\item dynamic masks: masks are generated anew whenever transition is sampled from replay buffer,
\item static masks: each transition is assigned a fixed mask generated when added to the replay buffer. 
\end{itemize}
In the cases of dynamic masks, each batch was split equally among the elements of ensemble (for this we kept the batch size to be the multiplicity of the number of ensembles). The static masks were inspired by the Bootstrapped DQN, see \citet[Appendix B]{OsbandBPR16}, where it is a core idea. We experimented with applying a different mask to each transition according to the Bernoulli distribution, or to assign the same masks for all transitions in the same trajectory.  

We found it useful to use static masks in the Deep-sea, Toy Montezuma's Revenge and Single-board Sokoban experiments, and dynamic masks in Multiple-board Sokoban experiments. 

\paragraph{Ensembles sampling} 
Recall that equation \ref{eq:action_choose} involves calculating the expected value, $\mathbb E_{\theta\sim\Theta}$, with respect to the posterior distribution $\Theta$.
In some experiments we instead sub-sample from $\Theta$. Such an approach follows \citet{OsbandBPR16}, which itself is inspired by the classical Thomson sampling (see discussion \citet[Section 4]{OsbandBPR16} and the original \citet{thomson}).  To be concrete, for a given a risk measure $\phi_a$, equation \ref{eq:action_choose} reduces to 
\begin{equation}\label{eq:subsampling}
\begin{split}
a^*(\mathtt{n})&:=\argmax_a \sum_{i\in\mathcal E}\left[\phi_a(\widehat{\mathbf{Q}}_{\theta_i}(\mathtt{n}))\right], \\ 
\widehat{\mathbf{Q}}_\theta(\mathtt{n}) &:= \left(\widehat{Q}_\theta(\mathtt{n},a')\colon a'\in\mathcal A\right),	
\end{split}
\end{equation}
where $\mathcal{E} = \{1, \ldots, K\}$. Sub-sampling, with a fixed parameter $\ell$, is equivalent to computing \eqref{eq:subsampling}, with $\mathcal{E}$ taken as a random subset of $\{1, \ldots, K\}$ of cardinality $\ell$.

\paragraph{Model training}
In this section, we describe the details of training of the environment model used for planning in the multi-board Sokoban experiments, see Section \ref{sec:sokoban}. As noted in the main text, we found out that randomly generated trajectories very rarely solve boards with $4$ boxes (in fact less than $3{e}{-6}$ of transitions contains solved boards). To ensure enough transitions resulting in a solved board, we sampled $100000$ episodes of length $40$ using boards containing $\{1,2, 3, 4\}$ boxes. In total, only $0.05\%$ of transitions were labeled as solved. To enable neural network training, we upsample positive transitions $100$ times. The model of the environment was not fine-tuned while training the RL agent; however, it was quite good. During this phase, we logged errors in model predictions encountered on the top-level trajectory (enrolled with the real environment). In $99.5\%$ of cases, the Sokoban frames predicted by the learned model matched the ground truth. Events of false reward predictions were even less frequent $\leq 1{e}{-6}$. However, we did not measure how many errors were made in the rollouts inside the search tree.

\paragraph{Randomized priors} 
It was shown in \citet[Lemma 3]{Osband2018a} that for Bayesian linear regression setting with Gaussian prior and noise model,   generating samples from posterior distribution is equivalent to solving an appropriate optimization problem. 
To be exact, suppose $\mathcal D=\{(x_i,y_i)\}$ is the dataset, $f_\theta(x)=x^T\theta$ is the regression function, $\epsilon_i$ is a Gaussian noise,  
$\tilde{\theta}$ comes from a Gaussian prior, and $\tilde{y}_i=y_i+\epsilon_i$. Then the solution of the following problem 
\begin{equation}\label{eq:random_priors}
\argmin_{\theta} \left(||\tilde{y}_i-(f_{\tilde{\theta}}+f_\theta)(x_i)||_2^2 + \zeta ||\theta||_2^2\right),
\end{equation}
for some $\zeta>0$,  enables one to sample from the posterior $\theta|\mathcal D$. 
Consequently, \citet{Osband2018a} propose to use \eqref{eq:random_priors} as a training objective and include randomized prior in the value function approximator. This objective is matched with the one used in Algorithm \ref{alg1}. We observed that the presence of randomized priors did not improve performance, hence we did not include them in final experiments. 

\section{Hyperparameters}
We summarize the parameters of the training used in Deep-sea, Toy Montezuma's Revenge, Single-board Sokoban and Multiple-boards Sokoban experiments presented in Section~\ref{sec:experiments}.

\begin{table*}[h]
\begin{center}
\begin{threeparttable}[b]
\begin{tabular}{lrrrr}\label{tab:ab}
Parameter  &  Deep-sea & Toy MR & Single-b. Sok. & Multiple-b. Sok. \\
\midrule
Number of MCTS passes\tnote{1}   &     $10$ & $10$ & $10$ & $10$ \\
Ensemble size $K$\tnote{2} &     $20$ & $20$ & $20$ & $3$ \\
Ensemble sub-sampling $\ell$\tnote{3}&     $10$ & $10$ & $10$ & no \\
Risk measure\tnote{4}& mean+std, & mean+std & mean+std & voting\\
$\kappa$\tnote{5}     &     $50$ &     $3$ &     $9$ &     n/a \\
VF target\tnote{6}    &    bootstrap &     bootstrap &     bootstrap &     bootstrap \\
Discounting $\gamma$\tnote{7}     &     $0.99$  &     $0.99$ &     $0.99$ &     $0.99$\\
Randomized prior\tnote{8}     & no    &  no   &  no    & no     \\
Optimizer\tnote{9}     &     RMSProp &     RMSProp &     RMSProp &     RMSProp \\
Learning rate\tnote{10}        &     $2.5{e}{-4}$ &   $2.5{e}{-4}$&     $2.5{e}{-4}$&     $2.5{e}{-4}$ \\

Batch size\tnote{11} &    32   &     32 &     32 &     32\\
Mask\tnote{12}     &    static   &     static &     static &     dynamic\\
\bottomrule
\end{tabular} 
\small{
\begin{tablenotes}
\item[1] $\mathtt{num\_mcts\_passes}$ in the MCTS algorithm, see Algorithm~\ref{alg:run_episode}
\item[2] the number of value functions in ensemble in Algorithm~\ref{alg1}
\item[3] the parameter of ensemble sub-sampling, see the second paragraph of Section \ref{sec:training_details}
\item[4] \textit{mean+std} stands for $\text{mean}+\kappa\cdot \text{std}$, where $\text{mean}$ is the mean of the ensemble predictions and $\text{std}$ is its standard deviation. \textit{voting} stands for using as given by \eqref{eq:voting}
\item[5] see footnote 4 
\item[6] see description in Section \ref{sec:method} and Algorithm \ref{alg:evaluate_episode} 
\item[7] as used in Algorithm~\ref{alg:evaluate_episode}
\item[8] see Section \ref{sec:method} and Appendix \ref{sec:training_details}
\item[9] the optimizer used in line $10$ of Algorithm \ref{alg1}
\item[10] the optimizer's learning rate
\item[11] cardinality of batch $B$ in line $9$  of Algorithm \ref{alg1}
\item[12] line $10$ of Algorithm \ref{alg1}
\end{tablenotes}
}
\end{threeparttable}
\end{center}
\caption{Default values of hyperparameters used in our experiments. }
\label{tab:hiperparameters}
\end{table*}

\section{Ablations}

In this section we analyze various experiment design choices. 

\subsection{Masks}
\begin{figure*}[h!]
\vskip 0.2in
\begin{center}
\centerline{\includegraphics[width=0.95\textwidth]{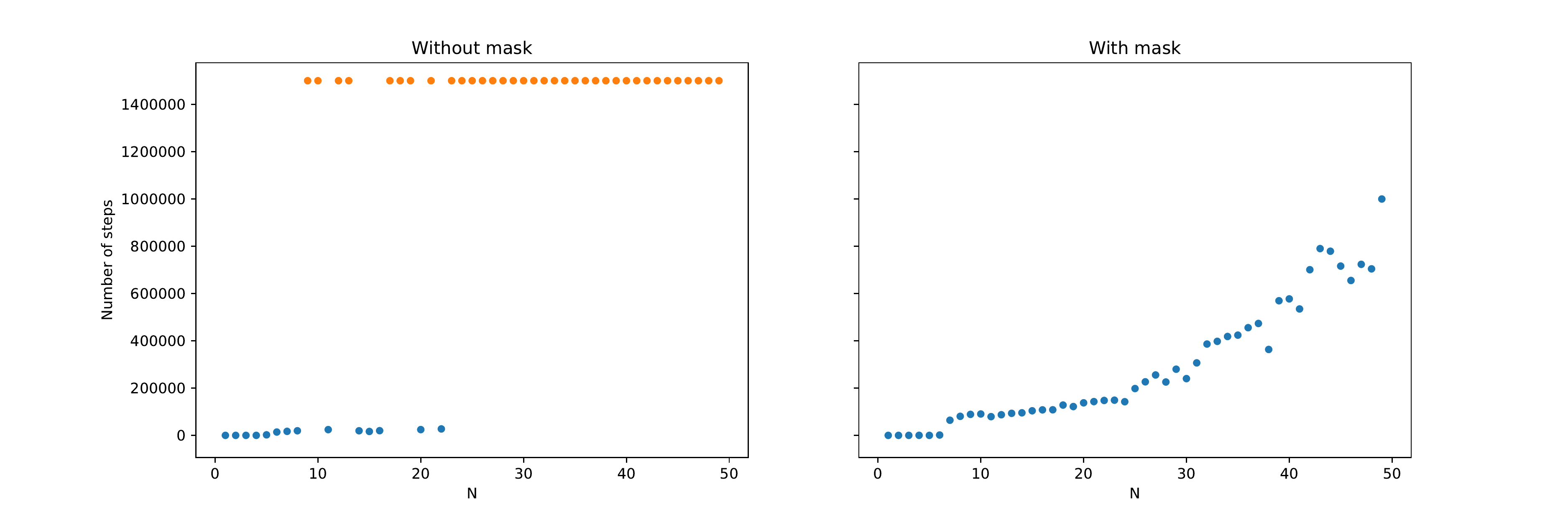}}
\caption{Deep-sea training with or without splitting training data between ensemble members.}
\label{fig:deep_sea_mask}
\end{center}
\vskip -0.2in
\end{figure*}

We conjecture that masks prevent premature convergence of  the members of the ensemble to the same values. We found that using masks (see Algorithm \ref{alg1} and Appendix \ref{sec:training_details}) was crucial in the Deep-sea environment. In Figure \ref{fig:deep_sea_mask} we observe failures of the setup without masks occurring on larger boards. Their careful analysis reveled that the agent often get stuck on a single, sub-optimal, trajectory. Masks also played important role also in the Toy Montezuma's Revenge environment, see Table \ref{table:montezuma-mask}. 

\begin{table}[h!]
\begin{center}
\begin{tabular}{ccc}\\\toprule  
Setup & solved / no. runs & av. visited rooms \\\midrule
\emph{without mask} &  22 / 50 & 13.0 \\  
\emph{with mask} & 30 / 37 & 17.5\\  \bottomrule
\end{tabular}
\caption{Toy Montezuma's Revenge - ratio of successful runs when learning ensembles with or without mask. Different runs were performed with different random seeds. }\label{table:montezuma-mask}
  \vspace{-10pt}
\end{center}
\end{table}

\subsection{Ensemble size}
\begin{figure*}[!htbp]
\vskip 0.2in
\begin{center}
\centerline{\includegraphics[width=0.95\textwidth]{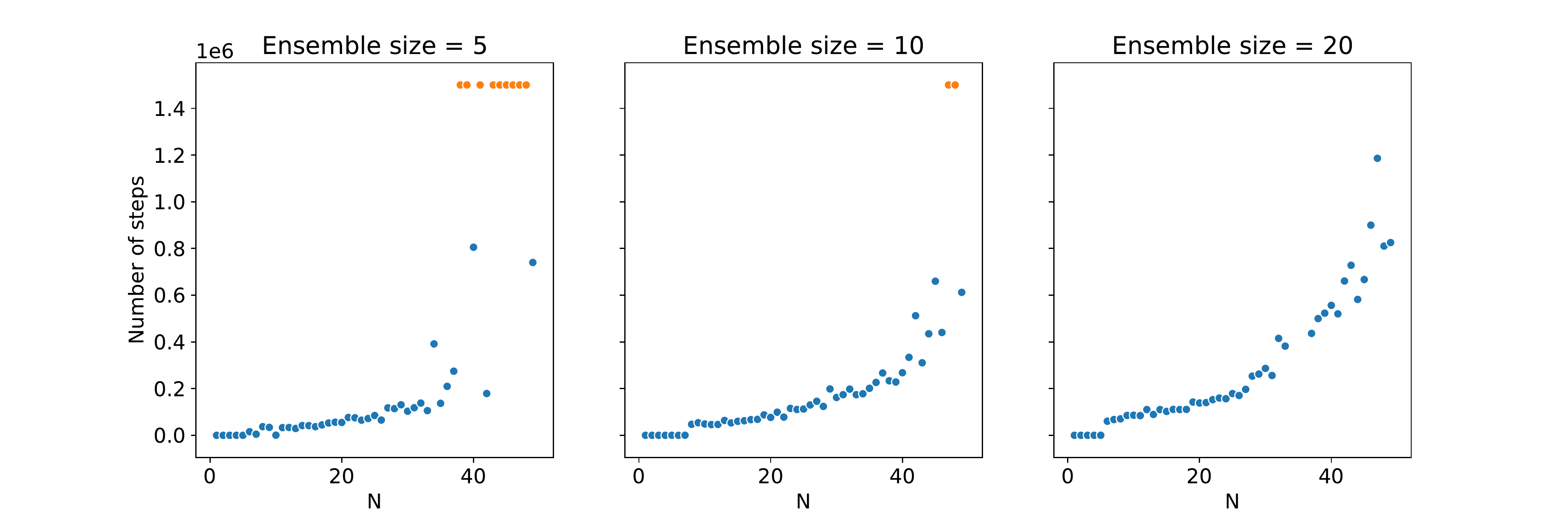}}
\caption{Deep Sea training with different ensemble sizes.}
\label{fig:deep_sea_ensemble_size}
\end{center}
\vskip -0.2in
\end{figure*}

\begin{figure}[!htbp]
  \begin{center}
    \includegraphics[width=1.\columnwidth]{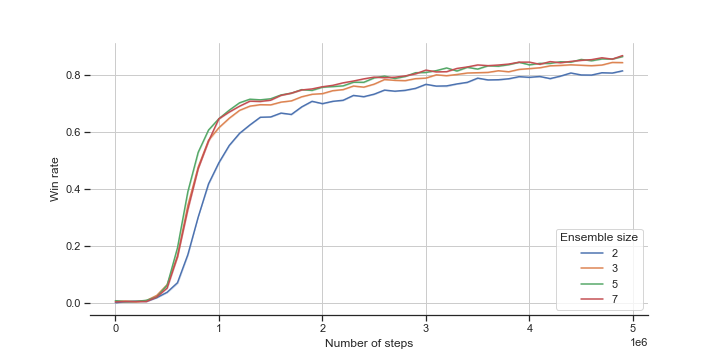}
    \caption{Multiple-boards Sokoban with different sizes of ensemble. Values 3, 5 and 7 result in similar performance.
    }\label{fig:multiboard_ensemble_size}
  \end{center}
\end{figure}

Using ensembles is crucial to our method as it enables us to define a risk-sensitive tree traversal policy, as indicated in \eqref{eq:action_choose} and Algorithm \ref{alg:choose_action}. The optimal number of ensembles seems to depend on the environment and probably other hyper-parameters. 
In Figure \ref{fig:deep_sea_ensemble_size} we compare the training behavior for different ensemble sizes for the Deep-sea (recall also Figure \ref{fig:deepsea}). Ensemble sizes as high as $20$ are needed in this case. 
For multi-board Sokoban experiments, see Section \ref{sec:sokoban}, we found out that the ensemble of size $3$ was enough to improve over no-ensemble baseline. Increasing the size of the ensemble offers only marginal further gains, see Figure \ref{fig:multiboard_ensemble_size}.

\subsection{Value target}\label{sec:valueTargetAblation}
\begin{figure*}[!htbp]
\vskip 0.2in
\begin{center}
\centerline{\includegraphics[width=0.95\textwidth]{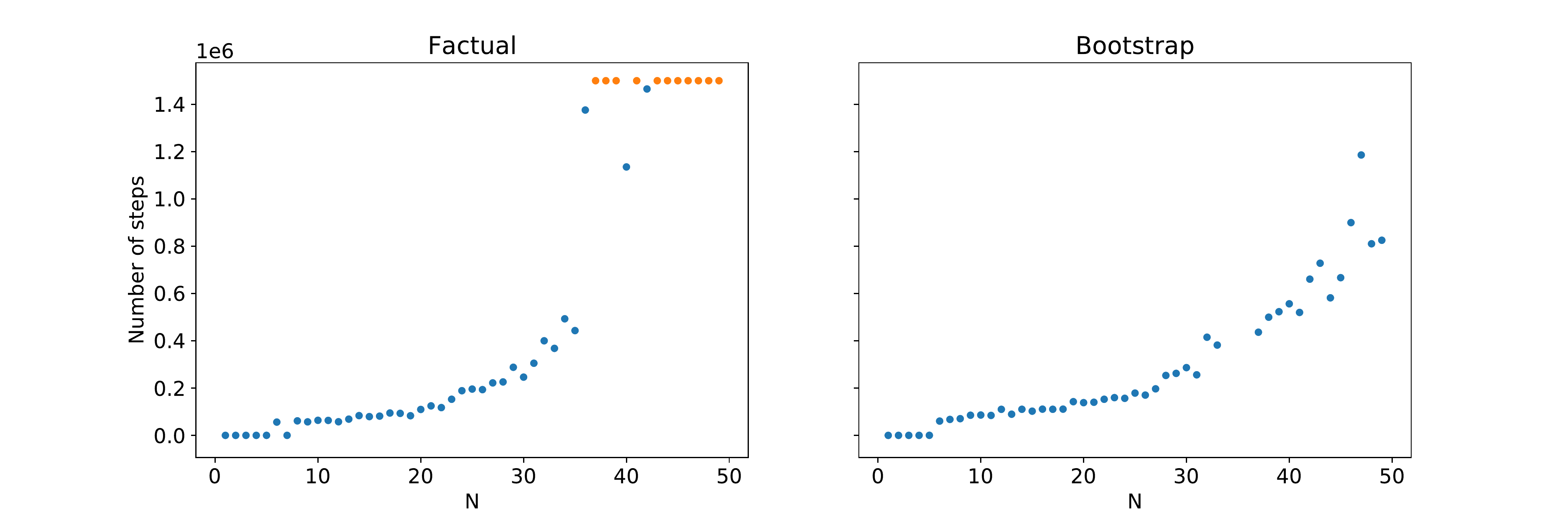}}
\caption{Deep-sea training with different training targets.}
\label{fig:deep_sea_value_target}
\end{center}
\vskip -0.2in
\end{figure*}

\begin{figure}[!htbp]
\vskip 0.2in
\begin{center}
\centerline{\includegraphics[width=\columnwidth]{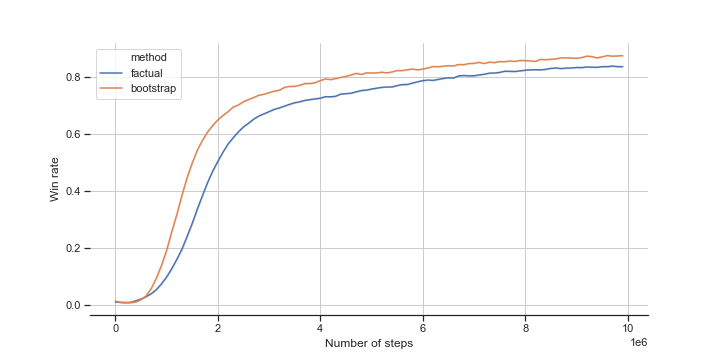}}
\caption{Multiple-boards Sokoban. Comparison of $\mathtt{factual}$ and $\mathtt{bootstrap}$ training targets.} 
\label{fig:multiboard_factual}
\end{center}
\vskip -0.2in
\end{figure}

Using $\mathtt{bootstrap}$, see Section \ref{sec:method} and Algorithm \ref{alg:evaluate_episode}, is a technical novelty of our work. We found that it works better than the standard $\mathtt{factual}$ method in all our experiments. In Figure \ref{fig:deep_sea_value_target} and Figure \ref{fig:multiboard_factual} we offer a comparison for the Deep-sea and Sokoban, respectively.

\subsection{Search size}
\begin{figure}[!htbp]
\vskip 0.2in
\begin{center}
\centerline{\includegraphics[width=\columnwidth]{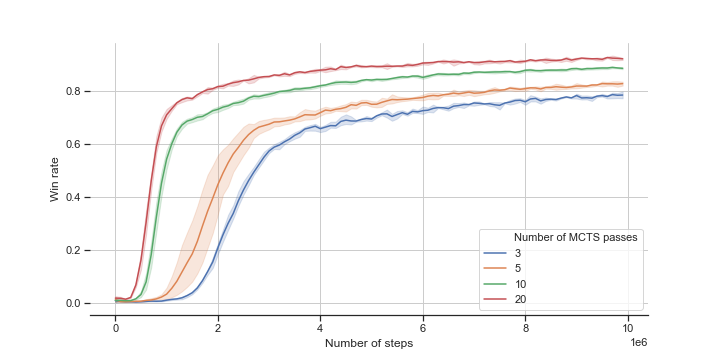}}
\caption{Multiple-boards Sokoban. Ensemble agent with different number of MCTS passes.} 
\label{fig:multiboard_mcts_passes}
\end{center}
\vskip -0.2in
\end{figure}

The parameter $\mathtt{num\_mcts\_passes}$ (see Algorithm \ref{alg:run_episode}) is perhaps the most important parameter of the planner. Increasing $\mathtt{num\_mcts\_passes}$ increases the computational costs and improves the quality of the planner. In this paper we purposefully focused on studies of exploration in the low computational-complexity regime setting $\mathtt{num\_mcts\_passes}=10$. This number is very modest compared to other works, for example in a recent paper \citet{Schrittwieser2019} declares $800$ passes was used to plan in board games. In Figure \ref{fig:multiboard_mcts_passes} we present a comparison of performance for various numbers of MCTS passes.

\begin{figure}[!htbp]
\vskip 0.2in
\begin{center}
\centerline{\includegraphics[width=\columnwidth]{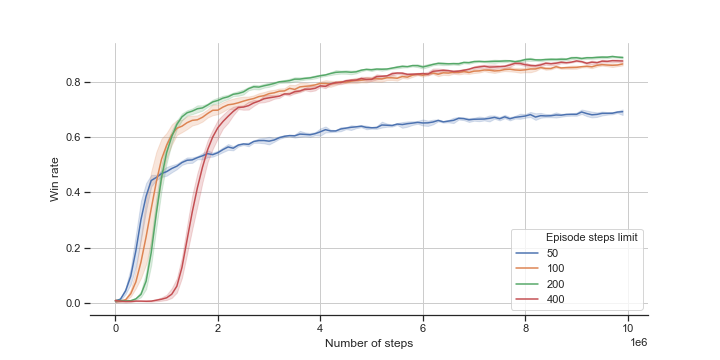}}
\caption{Multiple-boards Sokoban. Ensemble agent with different environment step limit per episode.}
\label{fig:multiboard_num_steps}
\end{center}
\vskip -0.2in
\end{figure}

Another important parameter is $\mathtt{max\_episode\_len}$. In the environments in which the episode does not have a fixed length, longer planning may yield positive results. In the multi-board Sokoban setting, 
increasing the values of $\mathtt{max\_episode\_len}$ slows down initial learning (weak agent will collect episodes of maximal length), 
but they ensure better learning at later phases of the training. We found $200$ to be a sweet-spot, which is $\approx 4$ times higher than the average solution length.  See Figure \ref{fig:multiboard_num_steps} for details.

\end{document}